\documentclass[letterpaper, 10 pt, conference]{ieeeconf}
\IEEEoverridecommandlockouts    
\usepackage{subcaption}
\usepackage{threeparttable}

\usepackage{graphics}           
\usepackage{times}              
\usepackage{amsmath}            
\usepackage{amssymb}            
\usepackage{graphicx}
\usepackage{algorithm}
\usepackage[noend]{algpseudocode}
\usepackage{booktabs}
\usepackage{color}
\usepackage[nocompress]{cite} 
\definecolor{instructioncolor}{rgb}{.5,.5,.5}
\definecolor{highlightcolor}{rgb}{0,0,1}

\usepackage[font=small]{caption}

\def\secref#1{Sec.~\ref{#1}}
\def\figref#1{Fig.~\ref{#1}}
\def\tabref#1{Tab.~\ref{#1}}
\def\eqref#1{Eq.~(\ref{#1})}

\makeatletter
\usepackage{xspace}
\DeclareRobustCommand\onedot{\futurelet\@let@token\@onedot}
\def\@onedot{\ifx\@let@token.\else.\null\fi\xspace}

\def\etal{{et al}\onedot}
\makeatother

\def\etalcite#1{\etal~\cite{#1}}

\usepackage{array}
\newcolumntype{L}[1]{>{\raggedright\let\newline\\\arraybackslash\hspace{0pt}}m{#1}}
\newcolumntype{C}[1]{>{\centering\let\newline\\\arraybackslash\hspace{0pt}}m{#1}}
\newcolumntype{R}[1]{>{\raggedleft\let\newline\\\arraybackslash\hspace{0pt}}m{#1}}

\usepackage{multirow}
\usepackage{siunitx}
\usepackage{url}

\usepackage{flushend}

\usepackage{tikz}
\newcommand\copyrighttext{%
  \footnotesize \textcopyright \the\year{} IEEE. Personal use of this material is permitted. Permission from IEEE must be obtained for all other uses, including reprinting/republishing this material for advertising or promotional purposes, collecting new collected works for resale or redistribution to servers or lists, or reuse of any copyrighted component of this work in other works.}
\newcommand\copyrightnotice{%
\begin{tikzpicture}[remember picture,overlay]
\node[anchor=south,yshift=10pt] at (current page.south) {\fbox{\parbox{\dimexpr0.75\textwidth-\fboxsep-\fboxrule\relax}{\copyrighttext}}};
\end{tikzpicture}%
}

\title{\LARGE \bf Leveraging GNSS and Onboard Visual Data from Consumer Vehicles for Robust Road Network Estimation}

\author{Balázs Opra \and
        Betty Le Dem \and
        Jeffrey M. Walls \and
        Dimitar Lukarski \and
        Cyrill Stachniss%
        \thanks{Balázs Opra, Betty Le Dem, Jeffrey M. Walls, and Dimitar Lukarski are with Woven by Toyota, Inc. Balázs Opra is also with the University of Bonn, Germany. Cyrill Stachniss is with the Center for Robotics, University of Bonn and with the Lamarr Institute for Machine Learning and Artificial Intelligence, Germany. 
        Email: \texttt{\{balazs.opra, betty.le.dem, jeff.walls, dimitar.lukarski\}@woven.toyota},  \texttt{cyrill.stachniss@igg.uni-bonn.de }}
}

\begin{document}
\maketitle
\copyrightnotice
\thispagestyle{empty}
\pagestyle{empty}

\begin{abstract}
    Maps are essential for diverse applications, such as vehicle navigation and autonomous robotics.
    Both require spatial models for effective route planning and localization.
    This paper addresses the challenge of road graph construction for autonomous vehicles.
    Despite recent advances, creating a road graph remains labor-intensive and has yet to achieve full
    automation. The goal of this paper is to generate such graphs automatically and accurately.
    Modern cars are equipped with onboard sensors used for today's advanced driver assistance systems like lane keeping.
    We propose using global navigation satellite system (GNSS) traces and basic image data acquired from these standard sensors in
    consumer vehicles to estimate road-level maps with minimal effort. We exploit the spatial information in the data by framing the problem
    as a road centerline semantic segmentation task using a convolutional neural network.
    We also utilize the data's time series nature to refine the neural network's output by using map matching.
    We implemented and evaluated our method using a fleet of real consumer vehicles, only
    using the deployed onboard sensors. Our evaluation demonstrates that our approach not only matches existing methods
    on simpler road configurations but also significantly outperforms them on more complex road geometries and topologies. This work received the 2023 Woven by Toyota Invention Award.
\end{abstract}

\section{Introduction}
\label{sec:intro}

Maps are foundational for vehicle navigation systems and indispensable for autonomous vehicles. They require
accurate road graphs for essential tasks like route planning, navigation, and localization. Such road graphs facilitate basic navigation and provide both geometric and semantic priors necessary for constructing advanced
lane-level High-Definition (HD) maps.
Despite progress in automated road graph extraction, especially from
GNSS traces, defined as sequences of geographic coordinates, achieving high-quality results remains a challenge
due to frequent inaccuracies that require substantial manual correction.

This paper addresses the problem of robust, high-quality road graph extraction. To tackle this, we introduce a unique
approach that utilizes both, GNSS traces and basic vision-based perception data, such as lane markings and road
boundaries, collected from consumer vehicles deployed today. We aim to substantially improve the quality of
automatically extracted road graphs, thereby reducing the manual workload traditionally required for their construction.

Road network inference has long been a research focus, leveraging various data sources, including but not limited to GNSS traces and aerial imagery.
Traditional methods have employed techniques such as k-means clustering and kernel density estimation~\cite{edelkamp2003csp, biagioni2012sigspatial}. With the advent of
deep learning, new approaches have emerged that use convolutional neural networks (CNNs) for semantic segmentation
and road graph inference~\cite{zhou2018cvprws, he2020eccv}. The challenge of accurately capturing complex road structures, especially those
with vertical dimensions like overpasses, remains when interpreting 2D input data. Our work proposes a new way to address such complexities.

\begin{figure}[t]
    \vspace{2.5mm}
    \centering
    \includegraphics[width=.49\linewidth]{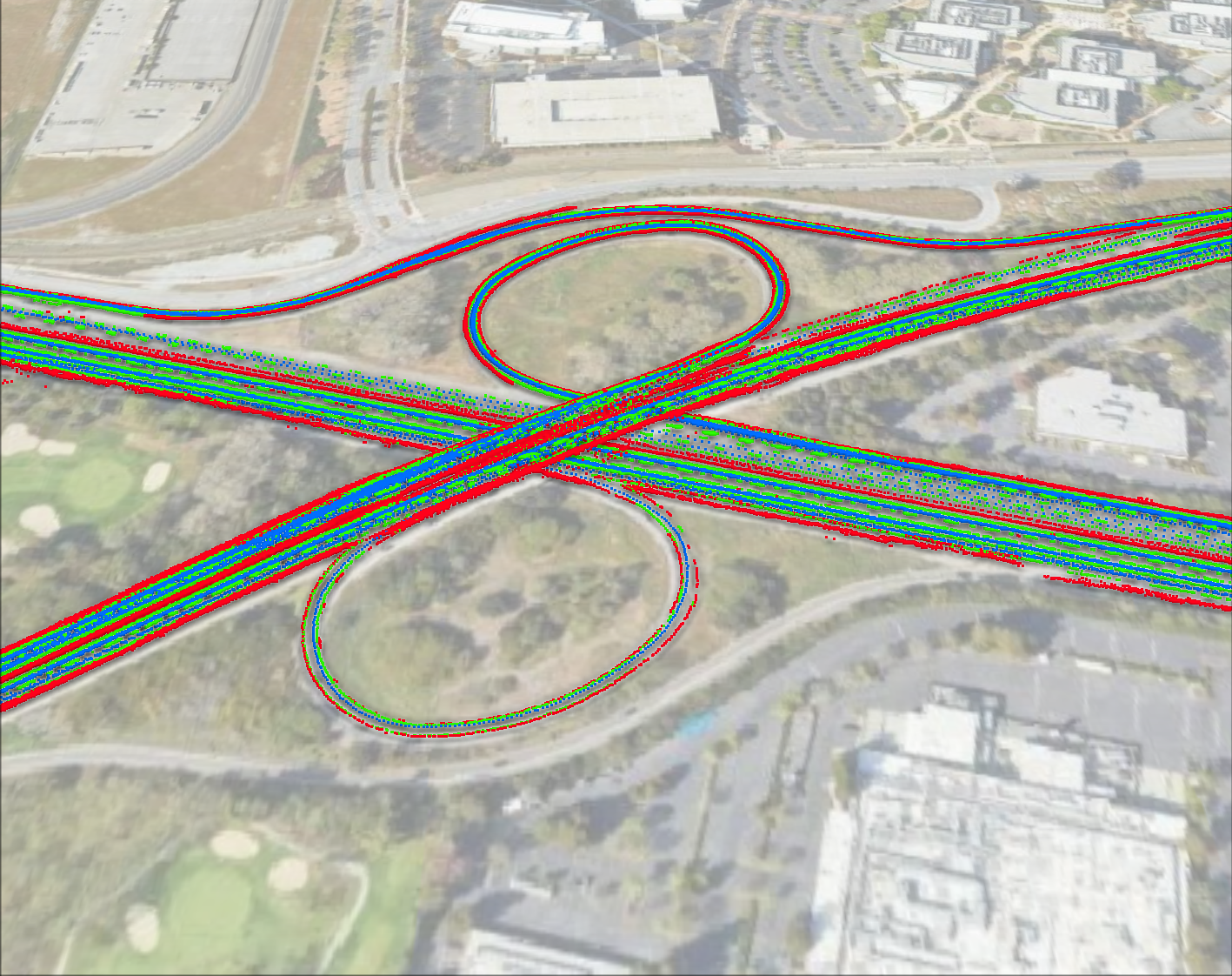}
    \includegraphics[width=.49\linewidth]{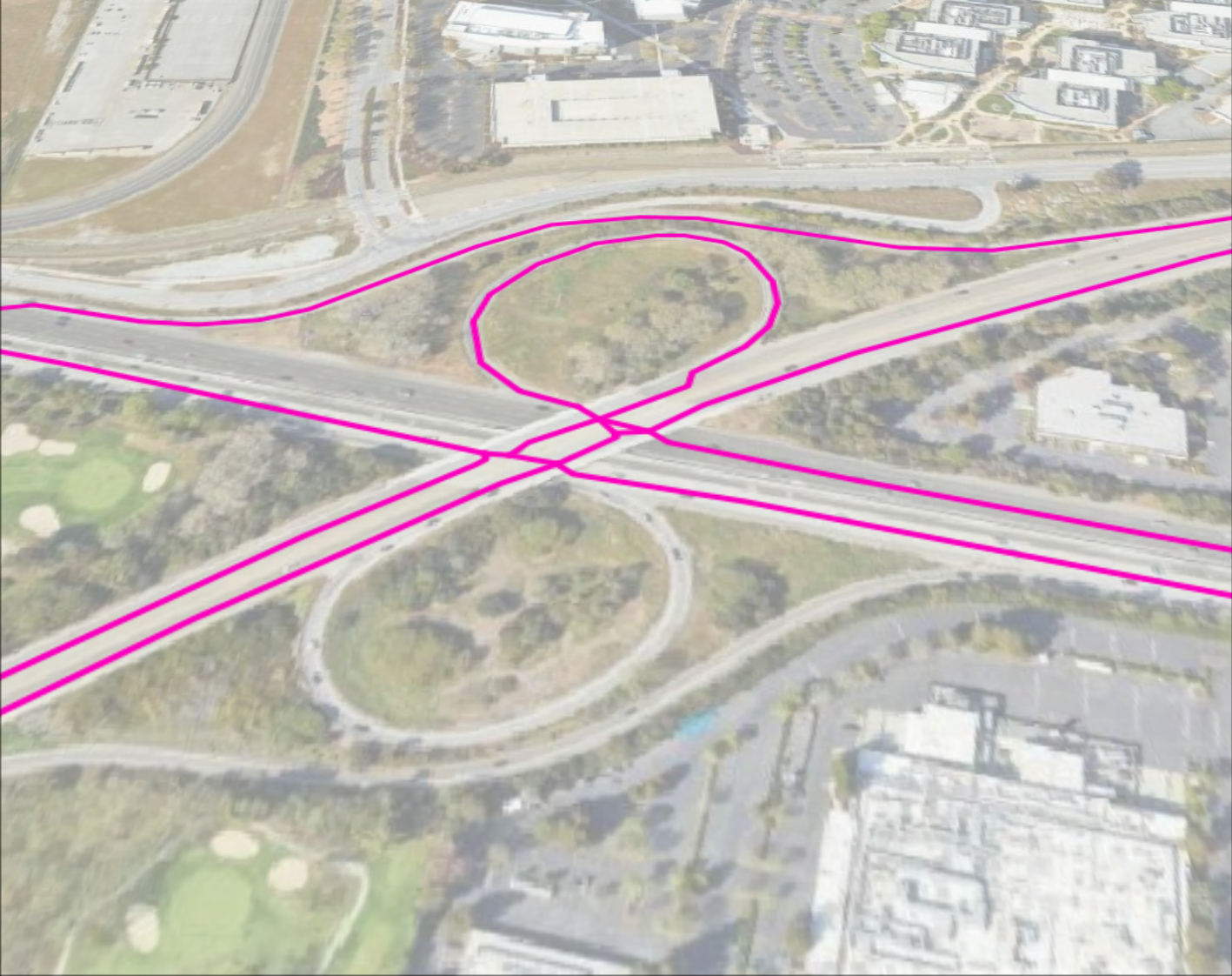}
    \caption{Left: Visualization of aligned fleet sensor data, red: road boundaries, green: lane markings, blue: GNSS trace points.  Right: Inferred road network of sufficiently covered roads while ignoring those driven only once.}
    \label{fig:motivation}
\end{figure}

The main contribution of this paper is a novel approach to robust and accurate road network inference. We utilize data from
vehicles equipped with standard consumer sensors. We only rely on standard GNSS data and information that
comes from Toyota's ADAS system which relies on a single camera plus odometry to extract lane markings and road boundaries. No extra data or custom sensor is used.
We furthermore employ a task-optimized deep learning framework to generate accurate road graphs. We optimize the network architecture and loss function specifically for road graph
inference. Our methodology includes novel ways to use map matching to curate the ground truth labels and
accurately detect stacked roads in the postprocessing steps. Besides achieving high-quality results, our methods show
strong generalization capabilities beyond the training area, all supported by rigorous experimental evaluation.
We make three key claims about our approach, namely that it can:
(i)~Infer highly accurate road graphs using a unique dataset of GNSS traces and features extracted from visual data from
standard consumer vehicles' sensors, all processed through an optimized deep-learning framework;
(ii)~Achieve robust generalization to unseen areas, including different countries, despite the limited size of our dataset.;
(iii)~Differentiate effectively between genuine road intersections and false positives such as bridges or underpasses,
using a novel map-matching-based postprocessing step.
The paper and our experimental evaluation back up these three claims.

\section{Related Work}
\label{sec:related}

Road graph inference has been approached using various data types, traditionally GNSS traces from vehicles or mobile phones. Early approaches include k-means clustering~\cite{edelkamp2003csp, schroedl2004dmkd},
kernel density estimation~\cite{davies2006pc, biagioni2012sigspatial}, and greedy iterative
graph construction~\cite{cao2009sigspatial,niehoefer2009spacomm}.
Other methods include trajectory clustering~\cite{buchin2017sigspatial, buchin2020sigspatial} and
iterative graph construction based on trace connectivity~\cite{he2018sigspatial}. These statistical and algorithmic methods often lack scalability due to limited generalizability and sensitivity to parameter settings.

The advancement of deep learning has enabled different input modalities to be used for road inference, the most common
one being orthographic imagery. We can categorize image-based approaches
into two main groups: semantic segmentation and graph inference. Methods focusing on semantic segmentation~\cite{zhang2018grsl,zhou2018cvprws} often use convolutional neural networks with encoder-decoder skip
connections and dilated convolutions to extract roads from the aerial images. Bandara \etalcite{bandara2021icra} propose
SPIN Road Mapper, which uses a spatial graph reasoning module in conjunction with CNNs. The Transformer architecture \cite{vaswani2017neurips},
after its success in natural language processing, has been successfully adapted to
image classification and semantic segmentation as well \cite{dosovitskiy2021iclr, chen2021arxiv-trans}, and thus is a
potential alternative to purely convolutional networks for road segmentation.

Graph inference approaches typically employ a CNN for semantic
segmentation or feature extraction and then infer the road graph from the computed features.
Máttyus \etalcite{mattyus2017iccv} directly infer the road graph from aerial images using deep learning for initial
segmentation and an algorithm to reason about missing connections as a shortest path problem. Other methods use graph
neural networks~\cite{bahl2022cvprws, chengkai2023igarss} or
encoding-dependent heuristics~\cite{he2020eccv, xie2022bmvc} to infer the road graph from the features computed by an
encoder network. Xu et al. propose various methods for image-based map feature extraction tasks, which aim to infer
topologically structured graphs, including road curb detection through imitation learning~\cite{xu2021ral}, road network
graph generation via Transformers
and imitation learning~\cite{xu2022tgars, xu2023ral}, and city-scale road boundary annotation with continuous graph
inference~\cite{xu2022ral}.
They also introduce CenterLineDet~\cite{xu2022icra}, which uses vehicle-mounted sensors and
a Transformer for lane centerline graph extraction.

Overall, the inference of road graphs from top-down images still poses significant challenges, including dealing with obstructions from
buildings, vegetation, and clouds. Additionally, the graph structure must be derived from two-dimensional input data, and this
becomes challenging for vertically stacked roads like highway overpasses.

Some of the above-mentioned methods have been applied to GNSS trace data
as well. Sun \etalcite{sun2018sigspatial, sun2019cvpr} combine a GNSS data-based heatmap with remote sensing imagery
to achieve better road segmentation results. Liu \etalcite{liu2022arxiv} propose a dual-enhancement module that leverages
the distinct information available in GNSS traces and aerial imagery for semantic segmentation, aiming to capitalize on the unique contributions of each data source.

To the best of our knowledge, Ruan \etalcite{ruan2020aaai} were the first to employ deep learning
using only GNSS traces. They extract features that capture both, the spatial characteristics and the connectivity between GNSS traces, and feed these features into a CNN for precise road centerline inference.
From our perspective, the most advanced method for GNSS trace data-based road graph inference to date is by Eftelioglu
\etalcite{eftelioglu2022bigspatial}. They employ rasterization techniques on variables such as average speed, bearing
distribution, and bearing change, in addition to trajectory density, to improve the quality of the resultant road graph. Our approach is similar to the above-mentioned methods
in that it uses rasterized trace data as input to a semantic segmentation network. However, we exploit additional
visual information available in customer vehicles today to build better maps.

Post-processing steps are often used to improve the quality of the road graph.
A commonly used technique is map matching, which aligns noisy GNSS coordinates with a predefined road network
using e.g. hidden Markov models~(HMM)~\cite{newson2009sigspatial} to address the inherent inaccuracies in GNSS
data. Biagioni \etalcite{biagioni2012sigspatial} and Ruan \etalcite{ruan2020aaai} employ map matching to filter
and refine road graph edges. Similarly, we use map matching for post-processing, most notably to disambiguate stacked roads in the road graph.

\section{Our Approach to Road Graph Inference}
\label{sec:main}

Our task is to infer the road-level street map in the form of a graph \( G = (V, E) \), where the vertices \( V \) represent
the location of intersections or dead-ends, and the edges \( E \) represent road centerlines. We use data retrieved from
vehicle fleets as input to our approach. This data contains GNSS traces \( S = \{ s_1, s_2, \ldots, s_n \} \),
each trace \( s_i \) comprising a sequence of three-dimensional geospatial coordinates \( s_i = (p_1, p_2, \ldots, p_k) \). The GNSS traces, fused from a single-antenna receiver, inertial sensors, and odometry, are relatively smooth and \emph{locally} accurate estimates due to the sensor fusion, and generally offer a meter-level absolute accuracy, which, however, may degrade significantly in urban environments.

The data associated with each trace also contains lane markings and road boundaries,
captured by the vehicle fleet's onboard camera-based detection system. This is Toyota's current
consumer vehicle detection system for driver assistance.
It consists of a forward-facing monocular camera and computation hardware that performs semantic segmentation and visual odometry in real-time.
The system outputs a sparse semantic point cloud, where each point is labeled as either a lane marking or a road boundary.
The data captured from a single vehicle is fairly sparse, e.g., a single dashed lane marking may only be represented by 5-10 points. No further data or sensors are used.
While our method uses this specific data, the approach can be adapted to any combination of point clouds, GNSS trajectories, and semantic segmentation that provides lane markings and road boundaries.

In a preprocessing step to our method, we utilize an existing offline SLAM approach to aggregate the GNSS traces and semantic point clouds. The SLAM system is based on the well-known paradigm of incremental smoothing and mapping proposed by Dellaert~\etalcite{dellaert2006ijrr}. Besides using the GNSS measurements and odometry,
it associates features from the semantic point cloud across multiple frames to construct a
factor graph. It iteratively performs feature association to add new measurements to the factor graph,
and least-squares optimization to refine the vehicle poses. This process improves the
estimation substantially and is essential in ensuring strong performance.

We refer to the aligned set of features---GNSS traces and the semantic point cloud from the vision data---as ``fleet sensor data'', see \figref{fig:motivation} for a visualization. We utilize this data in two ways. We exploit the spatial
information by framing the problem as a road centerline semantic segmentation task using a CNN, and we use the data's time
series nature to refine the neural network's output by map matching.

\subsection{Road Centerline Segmentation}
\label{sec:road-centerline-segmentation}

The first task in our road graph extraction method consists of extracting a binary road centerline segmentation mask
using a CNN. We describe the details of input data preparation, network architecture, and loss function below.

\textbf{Rasterization}. The sensor data includes both, 3D driving trajectories and a semantic point cloud featuring
lane markings and road boundaries as distinct classes.
To convert this 3D data into images, we project it to the ground plane and rasterize it.
For the trajectories, we begin with a 2D raster grid initialized to zero and trace
each path using Bresenham's algorithm~\cite{bresenham1965ibmsj}, incrementing a counter stored in each traversed pixel by 1. Similarly, for lane
markings and road boundaries, we start with zero-initialized rasters and increment the value of the raster pixel
corresponding to each point.

This produces three grayscale density images, each representing a specific semantic category. In our
implementation, we use raster grids with a resolution of \SI{1}{\meter} per pixel and a tile size of \( 1000 \times 1000 \) pixels.
We also conducted experiments with a higher resolution of \SI{0.2}{\meter} per pixel, however, we
found that the increased resolution does not lead to an improved road centerline inference performance but
comes at a significant increase in computational cost.
It seems that the \SI{1}{\meter} per pixel resolution is sufficient, probably as the road centerline is difficult to define unambiguously.

\begin{figure}[t]
    \centering
    \includegraphics[width=1\linewidth]{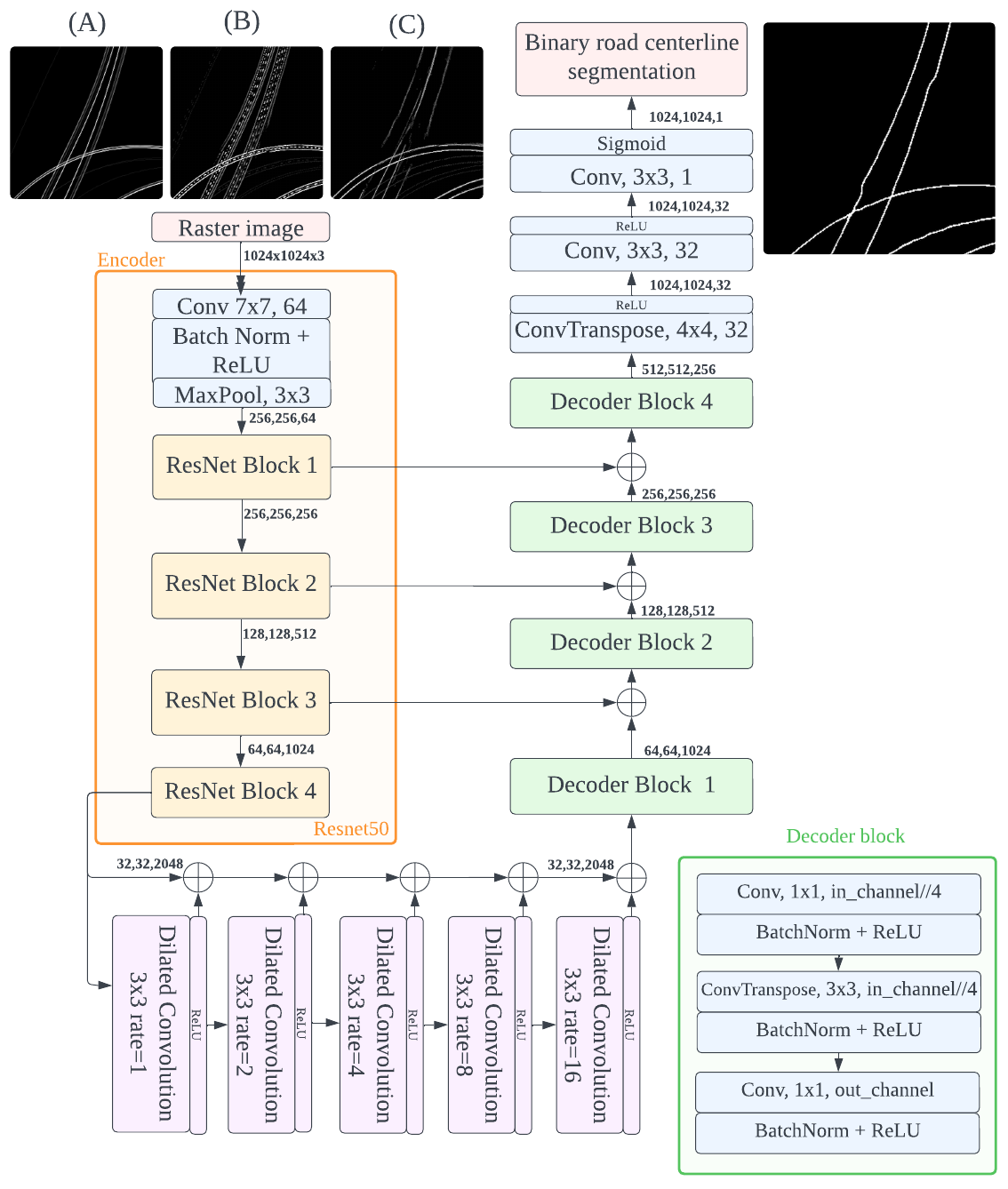}
    \vspace{-6mm}
    \caption{The D-LinkNet~\cite{zhou2018cvprws} model architecture used for road centerline segmentation from rasterized sensor data.
        Input data are the rasterized sensor data, namely traces (A), lane markers (B), and road boundaries (C).}
    \label{fig:model_architecture}
    \vspace{-3mm}
\end{figure}

\textbf{Network architecture.} We adopt the D-LinkNet architecture~\cite{zhou2018cvprws} to infer the road centerline
segmentation mask from our
unique rasterized sensor data. D-LinkNet was originally designed for high-resolution aerial image segmentation. It enhances the LinkNet~\cite{chaurasia2017vcip}
architecture by incorporating a central block with five dilated convolution layers and skip connections, as shown in
\figref{fig:model_architecture}. The large receptive fields enabled by the dilated convolution layers make it well-suited for
capturing long and thin roads.

The novelty in our work lies in adapting an existing network architecture, D-LinkNet, to a distinct data modality—--rasterized fleet sensor
data—--which includes features such as lane markings and road boundaries detected on board. While the original D-LinkNet uses pre-trained
ImageNet weights for its ResNet~\cite{he2016cvpr} encoder, we found that this approach resulted in poorer convergence for our data modality.
Consequently, we train the entire network from scratch. Additionally, our approach employs a different loss function tailored to our application, which we discuss below.

Since understanding road networks intuitively requires non-local spatial context, we experimented with incorporating Transformer-based architectures into our model.
We tried replacing the ResNet backbone with a
Bottleneck Transformer~\cite{srinivas2021arxiv}, and replacing the dilated convolution block with a Vision Transformer~\cite{dosovitskiy2021iclr}, similar to methods like TransUNet~\cite{chen2021arxiv-trans}.
Despite the potential of Transformers for capturing long-range dependencies, these modifications did not outperform D-LinkNet, likely due to our small dataset size and the challenge of segmenting narrow lines. This outcome suggests that convolutional networks are more suited for this particular task, benefiting from their inductive spatial bias and
data efficiency with small datasets.

\textbf{Loss function.} Our task is road centerline segmentation, where we want to predict pixels corresponding to the
road centerline skeleton as foreground. Properly segmenting the centerline is a challenging task as any
discontinuities negatively impact the quality of the resulting road graph. The authors of D-LinkNet
originally used the sum of the pixel-wise binary cross entropy (BCE) and the soft Dice loss as the loss function. This
function is well-suited for road segmentation where the foreground is typically thicker than a centerline, but it
often results in discontinuities when applied to the centerline segmentation task.

Thus, we optimize our network using the connectivity preserving loss (CP-loss) originally proposed by Xu
\etalcite{xu2021iros} to improve the continuity of road curb segmentation based
on aerial images. To realize better connectivity, we adapt the CP-loss to allow for a more flexible definition of the
road centerline. CP-loss is based on the combination of BCE and soft Dice losses, but increases the loss in areas where the skeletonized ground truth and predicted masks differ
outside a small buffer area around foreground pixels.
In our implementation, we approximate this buffer area by
applying a binary dilation to the skeletonized rasters with a \( 2 \times 2 \) structuring element. This effectively gives a margin of tolerance around the
ground truth and prediction skeletons, which is
desirable as it is ambiguous to specify the semantics of the road centerline with a sub-meter accuracy. We encourage the reader to
refer to the original work by Xu \etalcite{xu2021iros} for a more detailed explanation of the CP-loss.

\textbf{Label creation.} We use manually annotated road centerlines based on aerial images as labels for model training.
The annotations have complete road coverage in a given area, however, we never have uniform spatial coverage in our
sensor data. Some roads have a high number of trajectories, while others have only few to none. To address the sparsity of
sensor data coverage and be robust to potential erroneous outputs of the SLAM system, we
retain only those road edges in the ground truth that are traversed by a minimum number of GNSS traces. Removing roads
with too few traces enables the model to ignore areas with low data density, which may contain noise.
To determine which trace trajectories correspond to which road edges, we use the OSRM map matching algorithm
\cite{luxen2011sigspatial}, which is based on hidden Markov models. We only keep road edges that have at least $N$
trajectories associated with them. We found that $N=4$ gives solid results in our experiments and eliminates most issues.

\subsection{Road Centerline Refinement}
\label{sec:road-centerline-refinement}

The segmentation network generates a binary mask for each tile, with foreground pixels representing the predicted road centerline.
Our subsequent goal is to transform this binary mask into a road network graph, followed by the removal of any
spurious artifacts using a sequence of rule-based steps described below.

\textbf{Tile merging.} To create a single road graph for a large area, we perform inference on tiles that overlap by 50\% and merge the resulting segmentation masks using a weighted average.
The weight values are 1.0 in the center of the tile and decrease linearly to 0.5 at the edges. This selection of weights is based on the observation that due to the convolutional nature of the network,
the center of the tile has the most accurate predictions.

\textbf{Skeletonization.} We use the thinning algorithm by Guo and Hall~\cite{guo1989cacm} to produce a single pixel-wide skeleton
of the binary mask. The skeletonization tends to produce artifacts
at intersections as it generally fails to collapse them into a single node and may create a fake edge instead. In comparison
to other skeletonization methods such as Zhang-Suen~\cite{zhang1984cacm}, we found the Guo-Hall algorithm produces a skeleton where the fake
intersection edges are typically shorter. This makes it more suitable for our task.

\textbf{Vectorization.} We extract lines from the skeleton and turn them into a geospatial graph.
We create a node at each intersection of the line geometries. Some of these
intersections may not correspond to actual intersections in the road network but are instead representations of
multiple vertically stacked roads, e.g., bridges and underpasses. This is caused by the 2D segmentation
mask that cannot represent the vertical dimension. We address this issue in the intersection disambiguation step.

\textbf{Gap filling.} We fill discontinuities in the skeleton by adding connecting edges opportunistically, utilizing the
gap-filling algorithm proposed by Ruan \etalcite{ruan2020aaai}. In short, we iterate over all dead-end nodes in the graph and
try connecting them with either the closest nearby dead-end node or with edges that intersect the straight extension
line starting from the dead-end node. We only add a new edge if it is shorter than a specific threshold, and for node-to-node
connections we also require that the new edge does not result in a turn sharper than 90 degrees.

Since the gap-filling process connects the road extremities to each other, it may produce false positive edges.
To remove these, we match the sensor data trajectories to the proposed graph using the map-matching algorithm
proposed by Biagioni \etalcite{biagioni2012sigspatial}. In contrast to Biagioni et al., we assign edge-to-edge
transition probabilities based on whether an edge is from the gap-filling process or the semantic segmentation
network. Formally, the probability $P_{e,v}$ of transitioning onto edge $e$ at node $v$ is calculated as:
\begin{equation}
    P_{e,v} = \frac{\alpha(e)}{\sum_{e' \in \Gamma(v)} \alpha(e')},
    \label{eq:transition_weight}
\end{equation}
where \( \alpha(e) \) is the edge weight factor for edge \( e \), which can be either \( \alpha_g \) for gap-filling edges
or \( \alpha_s \) for segmentation-based edges, and \( \Gamma(v) \) is the set of all edges connected to node $n$.
We use weights \( \alpha_g < \alpha_s \) in our experiments to give segmentation-based
edges a higher transition probability.
Intuitively, this means that we prefer to stay on the segmentation-based edges, but we are willing to transition to gap-filling edges
if there is no other option. We remove all gap-filling edges that do not have at least a certain number of trajectories associated with them.
We use the same threshold value here as in the label creation step of the segmentation network.

\textbf{Graph cleaning.} We collapse intersection nodes that are connected by a single
short edge to remove the false positive intersections created by the skeletonization.
Furthermore, we remove dead ends shorter than \SI{15}{\meter} from the graph as they typically correspond to noise in
the segmentation output.

\textbf{Intersection disambiguation.} Since the segmentation mask cannot represent the vertical dimension of the road network,
we have to determine which of the intersections in the graph correspond to actual intersections in the road network,
as opposed to bridges or underpasses without an actual connection.

To accurately represent intersections in a road network, we exploit the time series nature of the sensor data. Our
algorithm first map-matches GNSS trajectories to the graph and identifies allowed transitions between edges at each
node based on a minimum number of supporting traces.
By grouping edges according to these allowed transitions and splitting nodes when multiple groups exist, the
algorithm refines the graph topology to include only valid road connections, filtering out false
intersections like bridges and overpasses.

We start by matching the sensor data trajectories to the output graph from the previous step to
determine the sequence of edges they travel through.
We use the matching algorithm proposed by Biagioni \etalcite{biagioni2012sigspatial} in this stage, setting uniform
edge-to-edge transition probabilities.
After matching, each coordinate of a trajectory is either associated with an edge of the graph or is unmatched.
Formally, for each sequence of trace coordinates \( s_i = (p_1, p_2, \ldots, p_k) \), map matching gives
us the result \( \tau_i \):
\begin{equation}
    \tau_i = ( \phi_1, \phi_2, \ldots, \phi_k ), \phi_i \in E \cup \{\text{Null}\},
\end{equation}
where \( E \) is the set of all edges in the graph.

Utilizing the map-matching results, we ascertain the frequently traversed edge-to-edge transitions at each
intersection node, as well as identify those transitions that are notably absent in the observed trace trajectories.
Formally, for each~\( \tau_i \), we determine the pairs of adjacent edges \( (e_i, e_j) \), which the trajectory
traverses. For each edge pair, we accumulate the number of occurrences across all traces and only consider those
edge pairs valid, which occur at least $N$ times. We use the same threshold value \( N = 4 \) here as in the label
creation step.

Upon identifying the edge-to-edge transitions supported by the trace data, we categorize the commonly traversed
edges into disjoint sets for each intersection node. In other words, for each intersection node \( v \), we group the incident edges \( \Gamma(v) \) into \( n \) disjoint sets \( \{E_{v,i}\}_{i=1}^{n} \) so that
all edges in a set \( E_{v,i} \) are indirectly connected through one or more valid transition
pairs, and \( \bigcup_{i}^{n} E_{v,i} = \Gamma(v),\,E_{v,i} \cap E_{v,j} = \emptyset \) for \( i \neq j \).

Should multiple disjoint edge sets exist for a given intersection node, this indicates that the location should not
be represented as a single intersection within the graph, but rather as distinct nodes.
Thus, if there are multiple disjoint sets for a node \( v \), we replace that node with multiple nodes
\( v'_{1}, \ldots, v'_{n}; \, n = | \mathcal{E}_v | \), and connect the edges in each \( E_{v,i} \) to the corresponding \( v'_i \).
This splits the intersection into multiple intersections, each corresponding to a set of edges that are
connected by allowable transitions. As the last step, we remove all nodes that have only two incident edges by
merging them into a single edge.

\section{Experimental Evaluation}
\label{sec:exp}
The main focus of this work is to extract the road graph from aggregated sensor data comprising GNSS traces
and a semantic point cloud collected from a consumer vehicle fleet without any further sensor or customization.

Our experiments demonstrate our method's capabilities, supporting the following claims: Our approach can
(i) Infer highly accurate road graphs using a unique dataset of GNSS traces and visual data from vehicles equipped with
standard consumer sensors, all processed through an optimized deep-learning framework;
(ii) Achieve robust generalization to unseen areas, including different countries, despite the limited size of our dataset;
(iii) Differentiate effectively between true intersections and false positives such as bridges or underpasses,
using a novel map-matching-based post-processing step.
\subsection{Experimental Setup}

\begin{figure}[t]
    \vspace{4pt}
    \centering
    \begin{subfigure}{.512\linewidth}
        \centering
        \includegraphics[width=.98\linewidth]{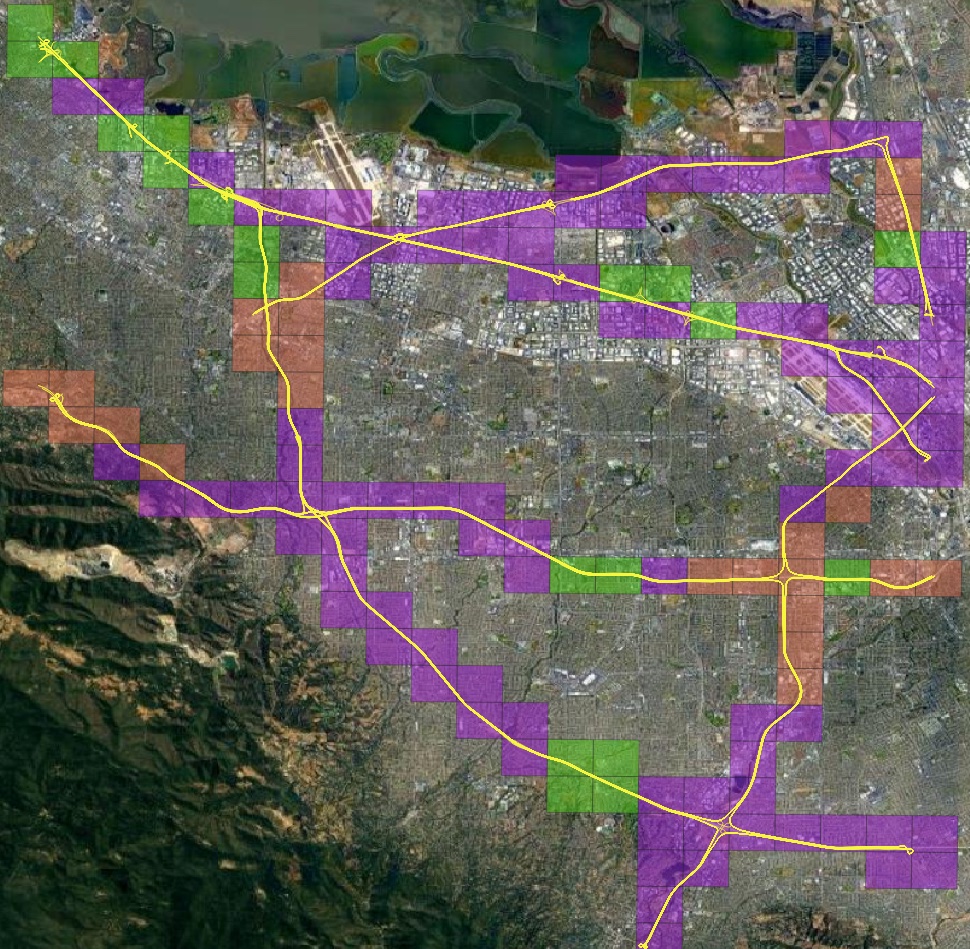}
    \end{subfigure}%
    \hfill
    \begin{subfigure}{.479\linewidth}
        \centering
        \includegraphics[width=.98\linewidth]{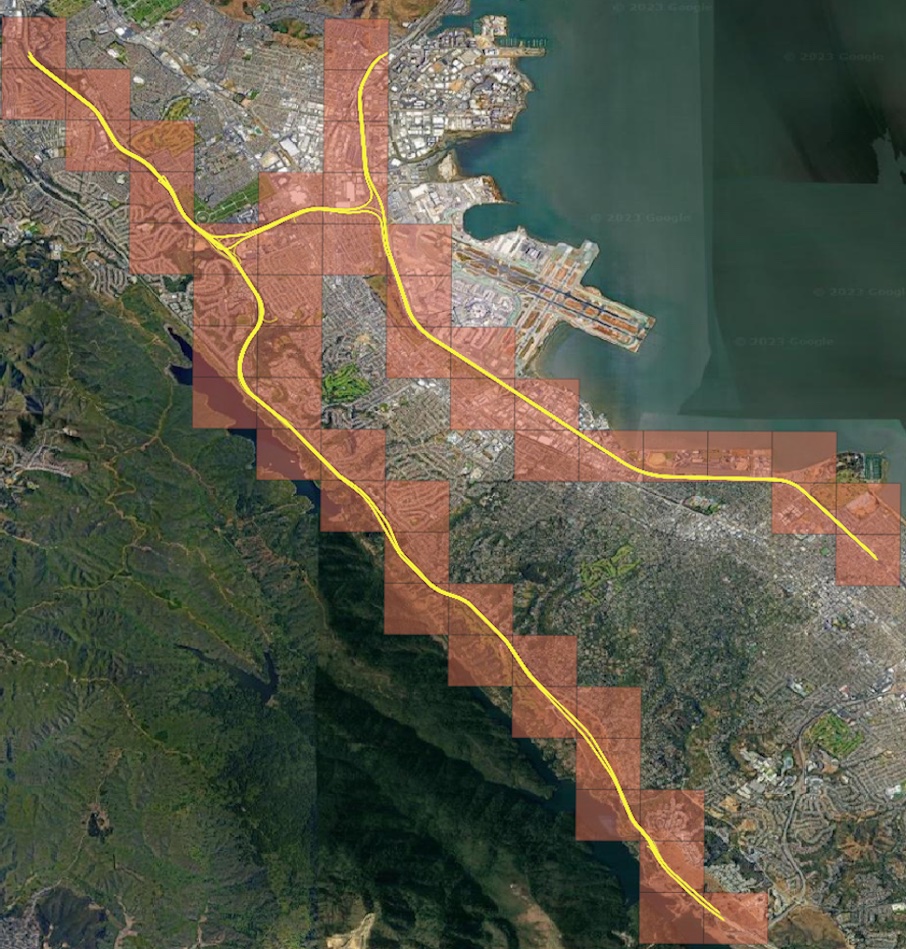}
    \end{subfigure}
    \\[0.9ex]
    \begin{subfigure}{\linewidth}
        \centering
        \includegraphics[width=.99\linewidth]{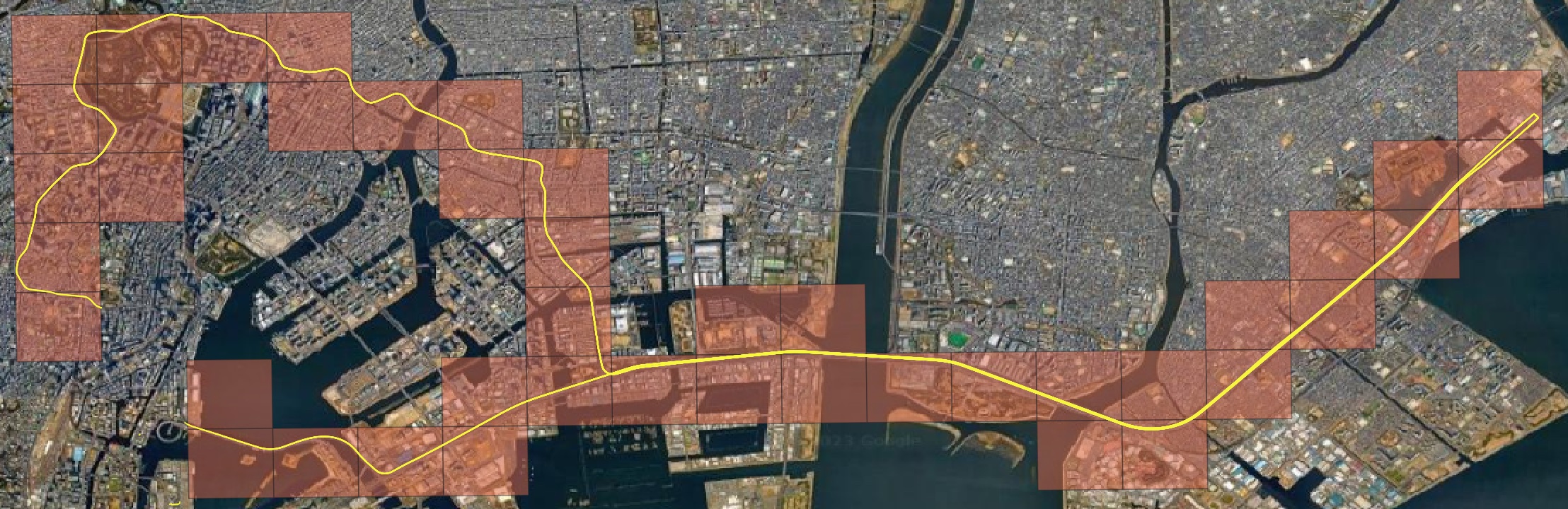}
    \end{subfigure}
    \caption{Dataset coverage used for the experiments. Top left: San Jose, top right: San Francisco, bottom: Tokyo.
        Aggregated GNSS traces are shown in yellow.
        Purple, green, and orange tiles represent training, validation, and test tiles respectively.}
    \label{fig:image2}
    \vspace{-3mm}
\end{figure}

\textbf{Dataset.} The proposed approach has been developed and validated on three datasets created from our vehicle fleet equipped with the standard customer vehicle sensors. To train the machine learning model and fine-tune the post-processing parameters, we use a dataset collected from highways in San Jose, California, which is
composed of \SI{236}{\kilo\meter} of trajectories. We split this dataset into
training/validation/test with 110/16/12 tiles of \(\SI{1}{\kilo\meter} \times \SI{1}{\kilo\meter}\) each.
We validate our approach on two test datasets from San Francisco, California, and Tokyo, Japan, covering \SI{78}{\kilo\meter} and \SI{47}{\kilo\meter} of highways, respectively.
The San Francisco test set has complex road topologies, while the Tokyo dataset presents a simpler setting, see \figref{fig:image2} for an overview.

\textbf{Implementation Details.}
We train the D-LinkNet-50 architecture from scratch for 500 epochs on a single NVIDIA Tesla V100 GPU, using the Adam optimizer~\cite{kingma2015iclr} and a constant learning rate of \(2 \times 10^{-4}\).
We found that more sophisticated learning rate schedules such as cosine annealing~\cite{smith2019aiml} lead to
overfitting and worse generalization, likely due to the limited size of our dataset.
A batch size of 2 yields the best results. We only use random rotation for data augmentation.
To normalize the input data, we apply a logarithmic transformation of \( \log_{10}(x + 1) \) to all pixels of all channels.
For the CP-loss, we set the value of the \( \sigma \) scaling parameter to 100.

\begin{table*}[t]
    \centering
    \resizebox{0.91\textwidth}{!}{%
        \begin{minipage}{\textwidth}
            \centering
            \vspace{8pt}
            \caption{Comparison of graph inference metrics across US datasets and methods.
                The best and second-best results on the San Francisco test dataset only are highlighted in bold and italics, respectively.
                For all metrics, higher is better.}
            \vspace{-1mm}
            \begin{tabular}{C{3.0cm}C{4cm}cccccc}
                \toprule
                                                      &                   & \multicolumn{3}{c}{GEO} & \multicolumn{3}{c}{iTOPO}                                                                     \\
                Dataset                               & Method            & Prec.                   & Rec.                      & \textbf{F1}    & Prec.          & Rec.           & \textbf{F1}    \\
                \midrule
                San Jose - Train                      &                   & 0.991                   & 0.968                     & 0.980          & 0.906          & 0.794          & 0.847          \\
                San Jose - Validation                 & Probe2Road (ours) & 0.959                   & 0.941                     & 0.950          & 0.787          & 0.778          & 0.783          \\
                San Jose - Test                       &                   & 0.949                   & 0.938                     & 0.943          & 0.7592         & 0.734          & 0.746          \\
                \midrule
                                                      & Woven baseline    & \textit{0.698}          & 0.692                     & \textit{0.695} & \textit{0.641} & 0.584          & 0.611          \\
                \multirow{2}{*}{San Francisco - Test} & Biagioni et al.   & 0.464                   & \textit{0.828}            & 0.595          & 0.388          & \textit{0.799} & 0.5219         \\
                                                      & DeepMG            & 0.648                   & 0.67                      & 0.673          & 0.567          & \textbf{0.883} & \textit{0.691} \\
                                                      & Probe2Road (ours) & \textbf{0.962}          & \textbf{0.935}            & \textbf{0.948} & \textbf{0.844} & 0.729          & \textbf{0.783} \\

                \bottomrule
            \end{tabular}
            \label{tab:graph_metrics}
        \end{minipage}
    }
    \vspace{-3mm}
\end{table*}

\begin{figure*}[t!]
    \centering
    \small
    \includegraphics[width=0.8515\linewidth]{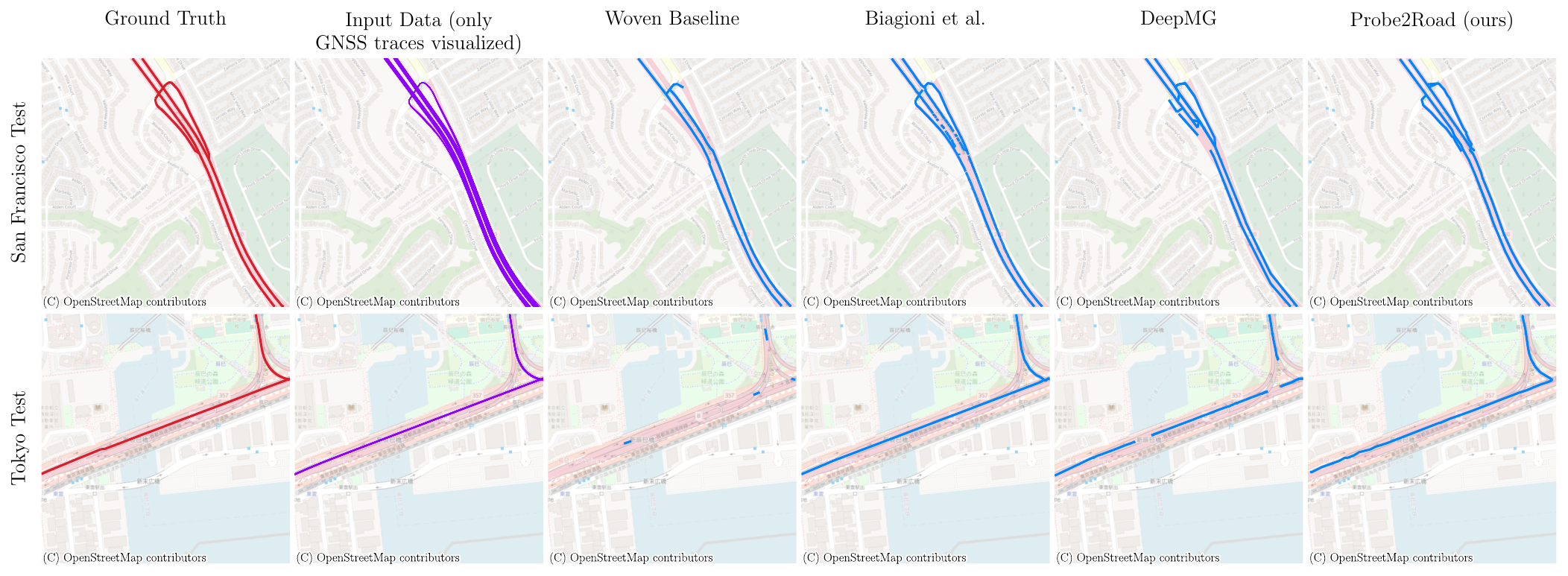}
    \vspace{-2mm}
    \caption{Results from all methods on selected areas of the San Francisco and Tokyo test sets. OpenStreetMap backdrop provided for context.}
    \label{fig:results_comparison}
    \vspace{-3mm}
\end{figure*}

\textbf{Metrics.} We use the following metrics to evaluate the performance of our approach and compare it with
baseline methods.
When evaluating the segmentation model, to account for the semantic flexibility of the road centerline—--which does not
need to be strictly aligned to the geometric center of the road—--we introduce what we call a "Soft F1 score."
Specifically, we apply binary dilation to both the ground truth and prediction masks using a \( 2 \times 2 \)
structuring element before calculating precision and recall. This dilation introduces a one-pixel margin around the
masks, allowing for a slightly more tolerant evaluation of model performance.

We adopt the GEO metric proposed by He \etalcite{he2022wacv} to evaluate the geometric accuracy of the road centerline.
We add a vertex every \SI{11}{\meter} in both graphs and set the search radius to \SI{6}{\meter}.
GEO evaluates local geometric accuracy but does not penalize topological errors such as small
discontinuities or false positive intersections. To address this limitation, we propose the intersection topology (iTOPO)
metric, a modification of Junction TOPO metric of Liao \etalcite{liao2023arxiv}.
This metric computes GEO scores for all subgraphs around all nodes of both the proposal graph and ground truth graph.
The computation begins by finding connected subgraphs around all nodes of both the proposal graph \( G = (V, E) \) and ground truth graph \( \hat{G} = (\hat{V}, \hat{E}) \).
Specifically for each \( v \in V \), we find the closest edge \( \hat{e} \in \hat{E} \) within a given distance (of \SI{6}{\meter}), if any.
If an edge is found we then consider the pair of subgraphs \( ( H_v, \hat{H}_{\hat{e}} ) \) on \( G \) and \( \hat{G} \),
respectively, where all geometries in \( H_v \) and \( \hat{H}_{\hat{e}} \) can be reached from \( v \) and \( \hat{e} \) by
traversing the graph for a maximum of \SI{30}{\meter}. If no edge is found, we set \( \hat{H}_{\hat{e}} = \emptyset \).
We repeat the same process for all \( \hat{v} \in \hat{V} \), creating subgraph pairs \( ( H_e, \hat{H}_{\hat{v}} ) \) for each
node in the ground truth graph. After having created all subgraph pairs,
\( \{ (H_v, \hat{H}_{\hat{e}}) \mid v \in V \} \cup \{ (H_e, \hat{H}_{\hat{v}}) \mid \hat{v} \in \hat{V} \} \)
, we invoke the GEO metric on them to
compute the aggregate precision, recall, and F1-score. This metric penalizes discontinuities and incorrect intersections in the proposal
graph.

\subsection{Performance of Our Approach}

\textbf{Comparison to other methods.} The first experiment evaluates the graph inference performance of our approach, labeled as Probe2Road, and benchmarks it against three GNSS-based methods: a popular and frequently used algorithm by Biagioni \etalcite{biagioni2012sigspatial}, a more recent, learning-based method by Ruan \etalcite{ruan2020aaai} called "DeepMG", and a Woven-developed method that uses kernel density
estimation, previously used before our method. These methods do not use data from the ADAS camera. To the best of our knowledge, no other method exists that uses a combination of GNSS and semantic point cloud data to infer road graphs.

The kernel density estimation algorithm voxelizes the GNSS traces into a 3D grid, uses Gaussian smoothing to approximate the kernel density estimate,
and thresholds it to produce a binary grid. Finally, it skeletonizes the 3D binary grid to create a graph.
For evaluation, we use a grid search over the algorithm's parameters to maximize the GEO-F1 score on the San Jose training dataset.
We train DeepMG on the same training set as Probe2Road using the settings in the source code published by the authors.
In \tabref{tab:graph_metrics} we report the GEO and iTOPO metrics for all methods on US datasets.

As shown in the table, Probe2Road clearly outperforms the other methods on the complex San Francisco test dataset, which contains intersections and overpasses. This supports the claim that our approach infers highly accurate road geometries with high topological correctness.

\begin{table}[]
    \vspace{2mm}
    \caption{Inference performance on the Tokyo test dataset, a simple dataset where geometric
        methods tend to perform well.
        \centering
        \small
    }
    \vspace{-1mm}
    \begin{tabular}{ccccccc}

        \toprule
                        & \multicolumn{3}{c}{GEO} & \multicolumn{3}{c}{iTOPO}                                                                     \\
        Method          & Prec.                   & Rec.                      & \textbf{F1}    & Prec.          & Rec.           & \textbf{F1}    \\
        \midrule
        Woven baseline  & 0.928                   & 0.050                     & 0.094          & \textbf{0.968} & 0.158          & 0.271          \\
        Biagioni et al. & \textit{0.971}          & \textbf{0.968}            & \textbf{0.970} & 0.877          & \textit{0.778} & \textbf{0.825} \\
        DeepMG          & 0.625                   & 0.569                     & 0.596          & 0.652          & \textbf{0.822} & 0.727          \\
        Probe2Road      & \textbf{0.979}          & \textit{0.950}            & \it{0.964}     & \textit{0.925} & 0.654          & \textit{0.767} \\
        \bottomrule
    \end{tabular}
    \label{tab:generalization}
    \vspace{-2mm}
\end{table}

\textbf{Generalization Study.} The second experiment assesses the generalization capabilities of our approach Probe2Road,
demonstrating its robustness in adapting to previously unseen geographical regions.
The quantitative results from the US datasets, as presented in \tabref{tab:graph_metrics} and \tabref{tab:soft_f1_metrics},
highlight the method's strong generalization performance. It demonstrates robust geometric and topological correctness on the San Francisco test dataset, matching
the performance on the San Jose training dataset and demonstrating excellent generalization to unseen areas within the same region.

When subjected to the significantly different highway structures of the Tokyo test dataset, Probe2Road performs very similarly to the US datasets, as shown in \tabref{tab:generalization},
showing that our method can successfully generalize to road networks in different countries.
The Woven baseline kernel density estimation-based algorithm fails at generalizing to the different GNSS trace distribution observed in the Tokyo dataset, achieving very low F1 scores.
The learning-based baseline method, DeepMG, also fails to generalize to the Tokyo dataset, achieving lower scores than Probe2Road on both metrics.
The Biagioni \etalcite{biagioni2012sigspatial} algorithm, while doing worse than the other methods on the San Francisco test dataset, performs well on the Tokyo test set.
On simpler roads like those in the Tokyo dataset, the vision-based point cloud data is not strictly required for high performance.
In these rather simple settings, with few intersections, our method is performing similarly to the method of Biagioni et al., which is the best performer on this dataset.
See \figref{fig:results_comparison} for visual comparisons of the results on the San Francisco and Tokyo test sets.
We provide more high-resolution visualizations that are hard to display in a paper on an accompanying website: {\small \url{https://bazs.github.io/probe2road}}

\begin{table}[]
    \caption{Soft F1 semantic segmentation metrics on all datasets.}
    \centering
    \small
    \vspace{-1mm}
    \begin{tabular}{lccc}
        \toprule
        Dataset              & Prec. & Rec.  & \textbf{F1} \\
        \midrule
        San Jose - Train     & 0.992 & 0.999 & 0.995       \\
        San Jose - Val       & 0.870 & 0.968 & 0.916       \\
        San Jose - Test      & 0.789 & 0.931 & 0.854       \\
        \midrule
        San Francisco - Test & 0.721 & 0.901 & 0.801       \\
        \midrule
        Tokyo - Test         & 0.862 & 0.975 & 0.915       \\
        \bottomrule
    \end{tabular}
    \label{tab:soft_f1_metrics}
    \vspace{-3mm}
\end{table}

\textbf{Ablation study on vision data.}
To confirm that incorporating the information extracted from the camera into Probe2Road,
as we propose, leads to better performance, we train our machine learning model using single-channel input data containing rasterized GNSS traces only,
denoted as (A) in \figref{fig:model_architecture}, and compare its performance against the model
trained on all sensor data, with three channels (A), (B), and (C) as shown in the same figure.

\tabref{tab:vision_ablation} indicates the scores
for the GNSS-only and the GNSS + vision data models on the San Francisco test dataset. The results support our claim that
the camera-based semantic point cloud data enhances Probe2Road's performance.

\begin{table}[]
    \vspace{2mm}
    \caption{GNSS-only and GNSS + vision data model performance on the San Francisco test dataset.
        \centering
        \small
    }
    \vspace{-1mm}
    \begin{tabular}{ccccccc}

        \toprule
                    & \multicolumn{3}{c}{GEO} & \multicolumn{3}{c}{iTOPO}                                                                      \\
        Model       & Prec.                   & Rec.                      & \textbf{F1}    & Prec.          & Rec.            & \textbf{F1}    \\
        \midrule
        GNSS only   & 0.880                   & 0.920                     & 0.900          & 0.483          & \textbf{ 0.776} & 0.596          \\
        GNSS+Vision & \textbf{0.962}          & \textbf{0.935}            & \textbf{0.948} & \textbf{0.844} & 0.729           & \textbf{0.783} \\
        \bottomrule
    \end{tabular}
    \label{tab:vision_ablation}
    \vspace{-2mm}
\end{table}
\begin{table}
    \caption{Performance on the San Francisco test dataset with no postprocessing, gap-filling only, and gap-filling + intersection disambiguation.}
    \centering
    \vspace{-1mm}
    \renewcommand{\arraystretch}{1.2}
    \resizebox{\columnwidth}{!}{%
        \begin{tabular}{lccccccc}
            \toprule
                           & \multicolumn{3}{c}{GEO} & \multicolumn{3}{c}{iTOPO}                                                                         \\
                           & Prec.                   & Rec.                      & \textbf{F1}     & Prec.           & Rec.            & \textbf{F1}     \\
            \midrule
            No postproc.   & 0.9607                  & 0.9297                    & 0.9449          & \textbf{0.8859} & 0.6839          & 0.7719          \\
            Gap filling    & \textbf{0.9618}         & \textbf{0.9348}           & \textbf{0.9481} & 0.8426          & 0.7285          & 0.7814          \\
            Full postproc. & \textbf{0.9618}         & \textbf{0.9348}           & \textbf{0.9481} & 0.8445          & \textbf{0.7293} & \textbf{0.7826} \\
            \bottomrule
        \end{tabular}%
    }
    \label{tab:post_processing_comparison}
    \vspace{-3mm}
\end{table}

\textbf{Effectiveness of the post-processing.} Our final experiment evaluates the effectiveness of the gap-filling and intersection disambiguation post-processing steps described in
\secref{sec:road-centerline-refinement}. We perform three evaluations to show the effectiveness of gap-filling, and the combination of gap-filling and intersection disambiguation.

Gap-filling boosts GEO-F1 by 0.34\% and iTOPO-F1 by 1.23\%, affirming its geometric correctness, but slightly diminishes
iTOPO precision. Notably, intersection disambiguation universally improves all iTOPO scores, even if slightly,
confirming its effectiveness in enhancing graph topology.

\section{Conclusion}
\label{sec:conclusion}

In this paper, we presented a novel approach to road graph extraction.
Our approach operates on data composed of GNSS traces and semantic point
clouds collected with the consumer vehicle fleet's sensors, which are available on modern cars today and are
used for advanced driver assistance functions such as lane keeping.
We utilize deep learning and rule-based steps to infer accurate
road graphs where enough data is available, enabling robust inference on highways.
We evaluated our approach on multiple datasets, providing comparisons to other existing techniques.
Our method demonstrates high geometric and topological accuracy, robust generalization, and our post-processing steps improve the quality for complex road structures.
The approach described in this paper won the 2023 Woven by Toyota Invention Award.

\section*{Acknowledgments}
We thank Louis Wiesmann for fruitful discussions and James Biagioni for providing code and assistance for~\cite{biagioni2012sigspatial}.

\bibliographystyle{plain_abbrv}
\bibliography{glorified,new}

\end{document}